\documentclass[journal,twocolumn]{IEEEtran}

\usepackage[normalem]{ulem}
\usepackage{url}
\usepackage{setspace}
\usepackage{hyperref}
\usepackage{multirow}
\usepackage{tikz}
\usepackage[maxfloats=25]{morefloats}
\usetikzlibrary{arrows}
\usetikzlibrary{shapes}
\usepackage{makecell}

\usepackage[numbers,square,sort&compress]{natbib}

\DeclareGraphicsExtensions{.pdf,.jpeg,.png}
\usepackage[caption=false,font=footnotesize]{subfig}
\usepackage{color}
\usepackage{rotating}

\usepackage[cmex10]{amsmath}
\usepackage{array}

\begin{document}

\renewcommand{\arraystretch}{1.6}
\newcommand{\highlight}[1]{\textcolor{red}{#1}}
\newcommand{\comment}[1]{}

\definecolor{darkgreen}{rgb}{0,.75,0}

\graphicspath{{./figures/}}



\title{Learning Aerial Image Segmentation\\ from Online Maps}

\author{Pascal Kaiser,
        Jan~Dirk~Wegner,
        Aur\'{e}lien Lucchi,
	Martin Jaggi,
	Thomas Hofmann,
        and~Konrad~Schindler
\thanks{All authors are with ETH Zurich, 
8093 Zurich, Switzerland}}

\markboth{IEEE Transactions on Geoscience and Remote Sensing}%
{Pascal Kaiser \MakeLowercase{\textit{et al.}}: Learning city
structures from online maps}

\maketitle

\begin{abstract}
This study deals with semantic segmentation of high-resolution (aerial)
images where a semantic class label is assigned to each pixel via supervised
classification as a basis for automatic map generation.
Recently, deep convolutional neural networks (CNNs) have shown
impressive performance and have quickly become the de-facto standard
for semantic segmentation, with the added benefit that task-specific
feature design is no longer necessary.
However, a major downside of deep learning methods is that they are extremely
data-hungry, thus aggravating the perennial bottleneck of supervised
classification, to obtain enough annotated training data.
On the other hand, it has been observed that they are rather robust
against noise in the training labels.
This opens up the intriguing possibility to avoid annotating huge
amounts of training data, and instead train the classifier from
existing legacy data or crowd-sourced maps which can exhibit high levels of noise.
The question addressed in this paper is: can training with large-scale, publicly 
available labels replace a substantial part of the manual labeling effort and still achieve 
sufficient performance? Such data will inevitably contain a significant portion of errors, but in return virtually
unlimited quantities of it are available in larger parts of the world.
We adapt a state-of-the-art CNN architecture for semantic segmentation
of buildings and roads in aerial images, and compare its performance
when using different training data sets, ranging from manually labeled,
pixel-accurate ground truth of the same city to automatic training
data derived from \textit{OpenStreetMap} data from distant locations.
We report our results that indicate that satisfying performance can be obtained
with significantly less manual annotation effort, by exploiting noisy large-scale training data.
\end{abstract}

%
\IEEEpeerreviewmaketitle


\section{INTRODUCTION}\label{INTRODUCTION}

\IEEEPARstart{H}{uge} volumes of optical overhead imagery are captured
every day with airborne or spaceborne platforms, and that volume is
still growing. This ``data deluge'' makes manual interpretation
prohibitive, hence machine vision must be employed if we want
to make any use of the available data.
Perhaps the fundamental step of automatic mapping is to assign a
semantic class to each pixel, i.e.\ convert the raw data to a
semantically meaningful raster map (which can then be further
processed as appropriate with, e.g., vectorisation or map
generalisation techniques).
The most popular tool for that task is supervised machine
learning. Supervision with human-annotated training data is necessary
to inject the task-specific class definitions into the generic
statistical analysis.
In most cases, reference data for classifier training is generated
manually for each new project, which is a time-consuming and costly
process. Manual annotation must be repeated every time the task, the
geographic location, the sensor characteristics or the imaging
conditions change, hence the process scales poorly.
In this paper, we explore the trade-off between:
\begin{itemize}
\item
  pixel-accurate %

but small-scale ground truth available; and
\item
  less accurate reference data that is readily available in arbitrary
  quantities, at no cost.
\end{itemize}

\noindent
For our study, we make use of online map data from \textit{OpenStreetMap}~\citep{haklay2008,haklay2010,girres2010}
(OSM, \url{http://www.openstreetmap.org}) to automatically derive
weakly labeled training data for three classes, \textit{buildings}, \textit{roads}, and
\textit{background} (i.e. all others). 
This data is typically collected using two main sources: (i) volunteers collect OSM data either in situ with GPS trackers or by manually digitizing very high-resolution aerial or satellite images that have been donated, and (ii) national mapping agencies donate their data to OSM to make it available to a wider public.
Since OSM is generated by volunteers, our approach can be seen as a form of crowd-sourced data
annotation; but other existing map databases, e.g.\ legacy data within a mapping agency, could also be used. 

As image data for our study, we employ high-resolution RGB orthophotos
from \textit{Google Maps}\footnote{specifications of Google Maps data 
can be found at \url{https://support.google.com/mapcontentpartners/answer/144284?hl=en}},
since we could not easily get access to comparable amounts of other
high-resolution imagery ($>$100$\,$km$^2$ at $\approx\,$10$cm$ ground sampling distance (GSD)).

Clearly, this type of training data will be less accurate. Sources of
errors include co-registration errors, e.g., in our case OSM polygons
and Google images were independently geo-referenced; limitations of
the data format, e.g., OSM only has road centerlines and category, but
no road boundaries; temporal changes not depicted in outdated map or
image data; or simply sloppy annotations, not only because of a lack of
training or motivation, but also because the use cases of most OSM users
require not even meter-level accuracy.

Our study is driven by the following hypotheses:
\begin{itemize}
\item
  The sheer volume of training data can possibly compensate for the
  lower accuracy (if used with an appropriate, robust learning
  method).
\item
  The large variety present in very large training sets (e.g.,
  spanning multiple different cities) could potentially improve the
  classifier's ability to generalise to new, unseen locations.
\item
  Even if high-quality training data is available, the large volume of
  additional training data could potentially improve the
  classification.
\item
  If low-accuracy, large-scale training data helps, then it may also
  allow one to substitute a large portion of the manually annotated
  high-quality data.
\end{itemize}

We investigate these hypotheses when using deep convolutional
neural networks (CNNs). Deep networks are at present the
top-performing method for high-resolution semantic labelling
and are therefore the most appropriate choice for our study.%
\footnote{All top-performing methods on big benchmarks are CNN
  variants, both in generic computer vision, e.g., the \textit{Pascal
    VOC Challenge}, \url{http://host.robots.ox.ac.uk/pascal/VOC/}; and in
  remote sensing, e.g., the ISPRS semantic labeling challenge,
  \url{http://www2.isprs.org/commissions/comm3/wg4/semantic-labeling.html}}
At the same time they also fulfil the other requirements for our
study: they are data-hungry and robust to label
noise~\citep{wegner2016}. And they make manual feature design somewhat
obsolete: once training data is available, retraining for different
sensor types or imaging conditions is fully automatic, without
scene-specific user interaction such as feature definition or
preprocessing.
We adopt a variant of the fully convolution network (FCN)
\citep{long2015}, and explore the potential of combining end-to-end
trained deep networks with massive amounts of noisy OSM labels.
We evaluate the extreme variant of our approach, without any manual
labelling, on three major cities (Chicago, Paris, Zurich) with
different urban structures.
Since quantitative evaluations on these large datasets are limited by
the inaccuracy of the labels, which is also present in the test sets,
we also perform experiments for a smaller dataset from the city of Potsdam.
There, high-precision manually annotated ground truth is available, which allows us
to compare different levels of project-specific input, including the
baseline where only manually labelled training data is used, the
extreme case of only automatically generated training labels, and
variants in between.
We also assess the models' capabilities regarding generalisation and
transfer learning between unseen geographic locations.

We find in this study that training on noisy labels does work well, but only
with substantially larger training sets. Whereas with small training
sets ($\approx\,$2$\,$km$^2$) it does not reach the performance of
hand-labelled, pixel-accurate training data.
Moreover, even in the presence of high-quality training data, massive
OSM labels further improve the classifier, and hence can be used to
significantly reduce the manual labelling efforts.
According to our experiments, the differences are really due to the
training labels, since segmentation performance of OSM labels is
stable across different image sets of the same scene.

For practical reasons, our study is limited to buildings and roads,
which are available from OSM; and to RGB images from Google Maps,
subject to unknown radiometric manipulations.
We hope that similar studies will also be performed with the vast
archives of proprietary image and map data held by state mapping
authorities and commercial satellite providers.
Finally, this is a step in a journey that this will ultimately bring us closer 
to the utopian vision that a whole range of mapping tasks no longer 
need user input, but can be completely automated by the world wide web.


\section{Related work}
\label{sec:RELWORK}

There is a huge literature about semantic segmentation in remote
sensing.  A large part deals with rather low-resolution satellite
images, whereas our work in this paper deals with very high-resolution
aerial images (see~\citep{rottensteiner2014} for an overview).

Aerial data with a ground sampling distance $GSD\leq20$cm contains
rich details about urban objects such as roads, buildings, trees, and
cars, and is a standard source for urban mapping projects. Since urban
environments are designed by humans according to relatively stable
design constraints, early work attempted to construct object
descriptors via sets of rules, most prominently for building detection
in 2D~\citep{fua1987,mohan1989} or in
3D~\citep{herman1984,weidner1997,fischer1998}, and for road
extraction~\citep{fischler1981,stilla1995,steger1995}.
A general limitation of hierarchical rule systems, be they top-down or
bottom-up, is poor generalization across different city layouts. Hard
thresholds at early stages tend to delete information that can hardly
be recovered later, and hard-coded expert knowledge often misses
important evidence that is less obvious to the human observer.

Machine learning thus aims to learn classification rules directly from
the data.
As local evidence, conventional classifiers are fed with raw pixel
intensities, simple arithmetic combinations such as vegetation
indices, and different statistics or filter responses that describe
the local image texture \cite{Leung2001,Schmid2001,Shotton2009}. An
alternative is to pre-compute a large, redundant set of local features
for training and let a discriminative classifier (e.g., boosting,
random forest) select the optimal subset
\cite{Viola2001,Dollar2009BMVC,Froehlich2013,tokarczyk2015features}
for the task.

More global object knowledge that cannot be learned from local pixel
features can be introduced via probabilistic priors. Two related
probabilistic frameworks have been successfully applied to this task,
Marked Point Processes (MPP) and graphical models.
For example,~\citep{stoica2004,chaiCVPR2013} formulate MPPs that
explicitly model road network topologies while~\cite{ortner2007} use a
similar approach to extract building footprints. MPPs rely on object
primitives like lines or rectangles that are matched to the image data
by sampling. Even if data-driven~\cite{verdie2014}, such Monte-Carlo
sampling has high computational cost and does not always find good
configurations.
Graphical models provide similar modeling flexibility, but in general
also lead to hard optimization problems. For restricted cases (e.g.,
submodular objective functions) efficient optimisers exist. Although
there is a large body of literature that aims to tailor conditional random fields (CRF) for
object extraction in computer vision and remote sensing, relatively few
authors tackle semantic segmentation in urban
scenes~\citep[e.g.][]{kluckner2009,kluckner2010,wegner2013,montoya2014,wegner2015}.

Given the difficulty of modeling high-level correlations, much effort
has gone into improving the local evidence by finding more
discriminative object
features~\cite{herold2003spatial,dallamura2010,tokarczyk2015features}.
The resulting feature vectors are fed to a standard classifier (e.g.,
Decision Trees or Support Vector Machines) to infer probabilities per
object category. Some authors invest a lot of effort to reduce the
dimension of the feature space to a maximally discriminative
subset~\citep[e.g.][]{Schwartz2009,Hussain2010,vanCoillie2007,
  Rezaei2012}, although this seems to have only limited effect -- at
least with modern discriminative classifiers.

Deep neural networks do not require a separate feature definition
step, but instead learn the most discriminative feature set for a given
dataset and task directly from raw images.
They go back to
\cite{fukushima1980neocognitron,lecun1989backpropagation}, but at the
time were limited by a lack of compute power and training data.
After their comeback in the 2012 ImageNet
challenge~\citep{russakovsky2015, krizhevsky2012imagenet}, deep
learning approaches,
and in particular deep convolutional neural networks (CNNs), have
achieved impressive results for diverse image analysis
tasks. State-of-the-art network architectures~\citep[e.g.,][]{simonyan2015very} have many (often 10-20, but up to
$>$100) layers of local filters and thus large receptive fields in the
deep layers, which makes it possible to learn complex local-to-global
(non-linear) object representations and long-range contextual
relations directly from raw image data.
An important property of deep convolutional neural networks (CNN) is
that both training and inference are easily parallelizable, especially on GPUs, and thus
scale to millions of training and testing images.

Quickly, CNNs were also applied to semantic segmentation of
images~\citep{Farabet2013}.  Our approach in this paper is based on
the fully convolutional network (FCN) architecture
of~\citep{long2015}, which returns a structured, spatially explicit
label image (rather than a global image label).
While spatial aggregation is nevertheless required to represent
context, FCNs also include in-network upsampling back to the
resolution of the original image. They have already been successfully
applied to semantic segmentation of aerial images,
~\citep[e.g.,][]{paisitkriangkrai2015,lagrange2015benchmarking,marmanis2016}.
In fact, the top performers on the ISPRS semantic segmentation
benchmark all use CNNs. We note that (non-convolutional) deep networks
in conjunction with OSM labels have also been applied for patch-based
road extraction in overhead images of $\approx1m$ GSD at large
scale~\citep{Mnih2010,mnih2012}.
More recently,~\citep{mattyus2015} combine OpenStreetMap (OSM) data
with aerial images to augment maps with additional information from
imagery like road widths. They design a sophisticated random field to
probabilistically combine various sources of road evidence, for
instance cars, to estimate road widths at global scale using OSM and
aerial images.

To the best of our knowledge, only two works have made 
attempts to investigate how results of CNNs trained on large-scale OSM labels 
can be fine-tuned to achieve more accurate results for labeling 
remote sensing images~\citep{mnih2013,maggiori2017}. However, we are not aware of any large-scale, systematic, comparative and quantitative study that investigates using large-scale training labels from inaccurate map data for semantic segmentation of aerial images.


\section{METHODS}\label{sec:METHODS}

We first describe our straight-forward approach to generate training
data automatically from OSM, and then give technical details about the
employed FCN architecture and the training procedure used to train our
model.

\subsection{Generation of Training Data}\label{sec:methods_data}

We use a simple, automatic approach to generate datasets of very-high
resolution (VHR) aerial images in RGB format and corresponding labels
for classes \textit{building}, \textit{road}, and
\textit{background}. Aerial images are downloaded from Google Maps and
geographic coordinates of buildings and roads are downloaded from
OSM. We prefer to use OSM maps instead of Google Maps, because the
latter can only be downloaded as raster images\footnote{Note that some 
national mapping agencies also provide publicly available map and other geo-data, e.g. 
the USGS national map program:~\url{https://nationalmap.gov/}}. OSM data can be
accessed and manipulated in vector format, each object type comes with
meta data and identifiers that allow straight-forward filtering.
Regarding co-registration, we find that OSM and Google Maps
align relatively well, even though they have been acquired and
processed separately.\footnote{Note that it is technically possible to
  obtain world coordinates of objects in Google Maps and enter those
  into OSM, and this might in practice also be done to some
  extent. However, OSM explicitly asks users not to do that.}
Most local misalignments are caused by facades of high buildings that
overlap with roads or background due to perspective effects. It is
apparent that in our test areas Google provides ortho-photos rectified
w.r.t.\ a bare earth digital terrain model (DTM), not ``true'' ortho-photos rectified with a digital surface model (DSM).
According to our own measurements on a subset of the data, this effect is relatively mild, 
generally $<10$ pixels displacement. We
found that this does not introduce major errors as long as there are
no high-rise buildings. It may be more problematic for extreme scenes
such as Singapore or Manhattan.

\begin{figure*}
	\centering
    \begin{tabular}{@{}c@{\hspace{5mm}}c@{\hspace{5mm}}}
		\includegraphics[width=0.45\linewidth]{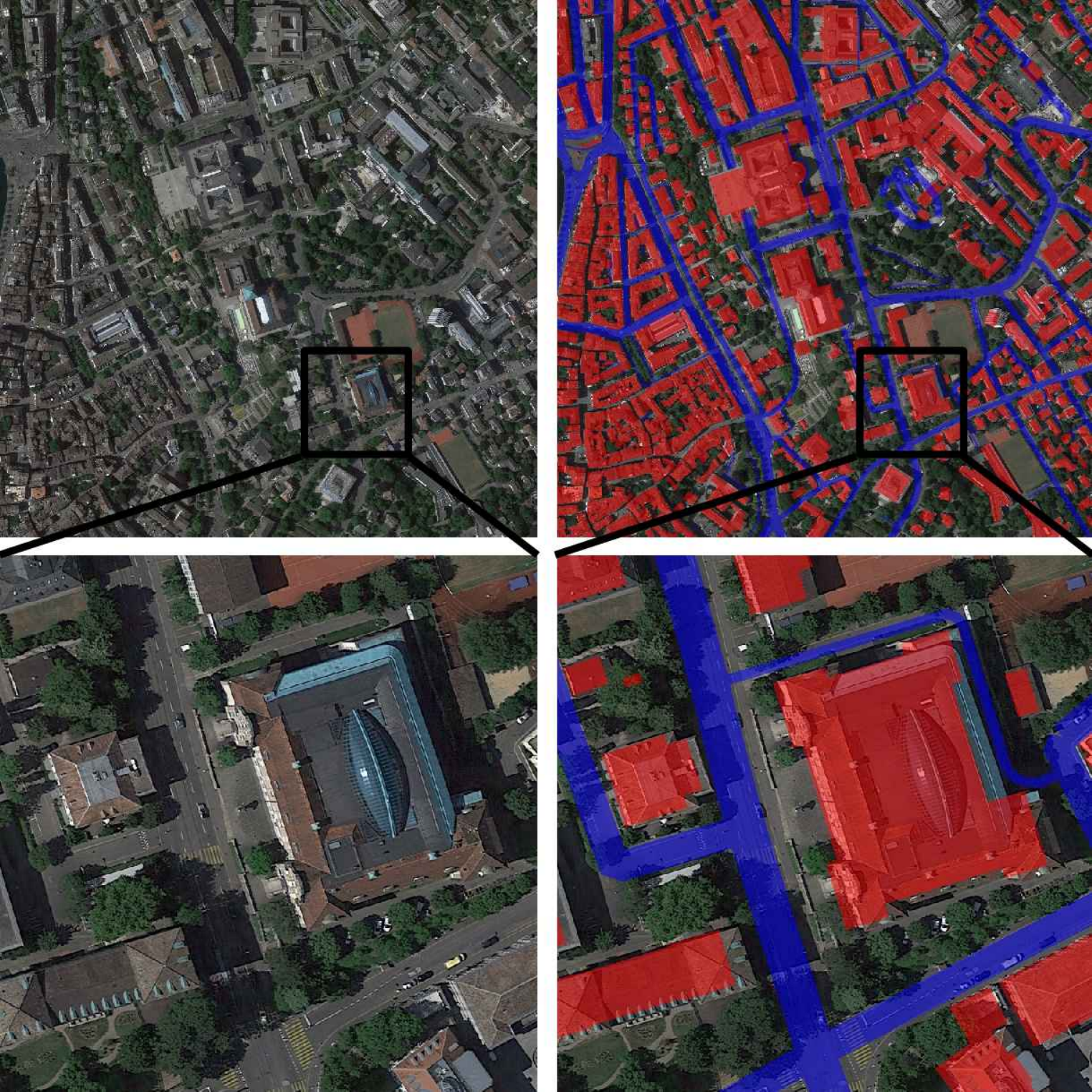} &
		\includegraphics[width=0.45\linewidth]{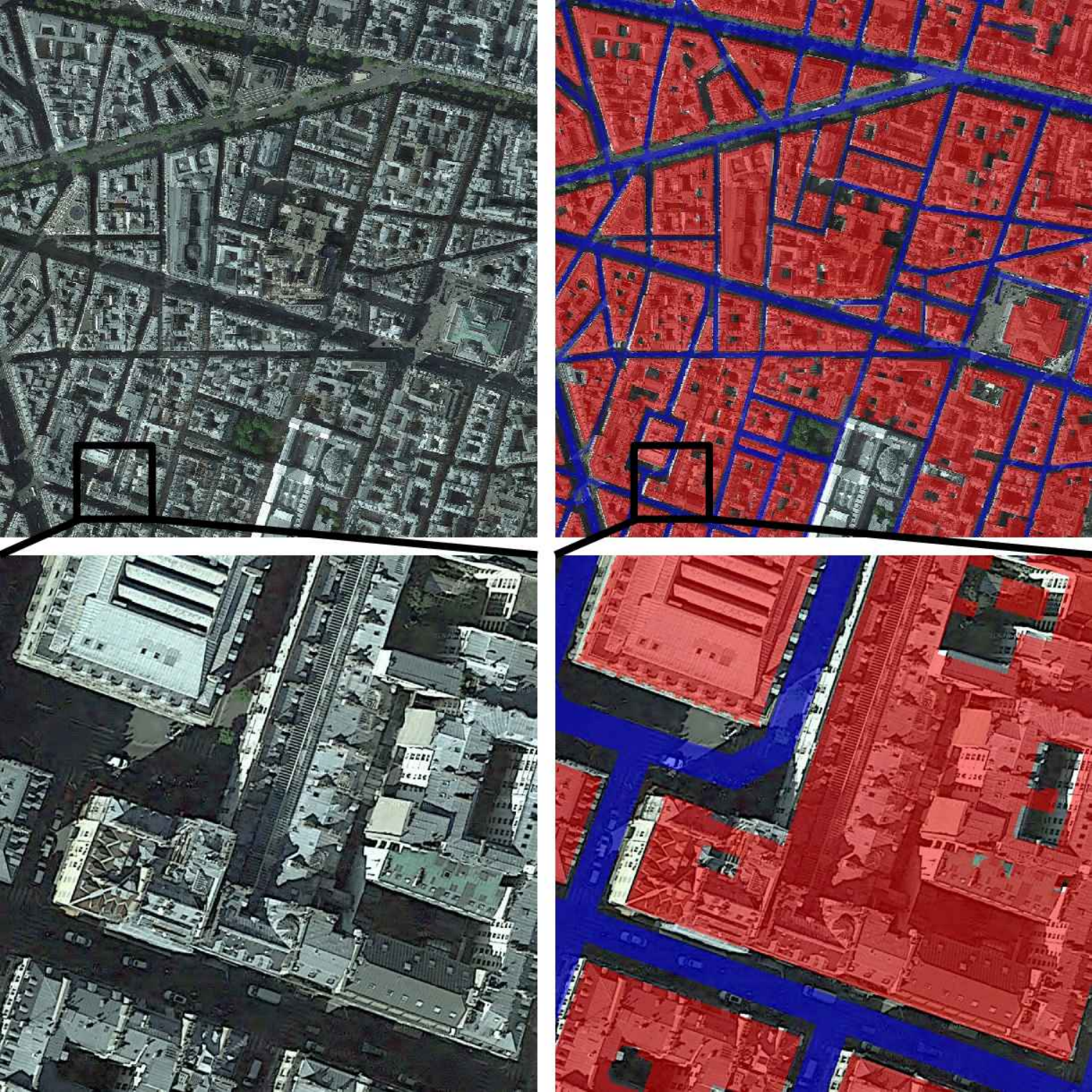}\\
    (a) & (b) \\
	  \end{tabular}
	\caption{Example of OSM labels overlaid with Google Maps
          images for (a) Zurich and (b) Paris. The left side shows an
          aerial image and a magnified detail. The right side shows
          the same images overlaid with building (red) and road (blue)
          labels. Background is transparent in the label map.}
\label{fig:img_ground_truth}
\end{figure*}

To generate pixel-wise label maps, the geographic coordinates of OSM
building corners and road center-lines are transformed to pixel
coordinates. For each building, a polygon through the corner points is
plotted at the corresponding image location.
For roads the situation is slightly more complex. OSM only provides
coordinates of road center-lines, but no precise road widths. There
is, however, a road category label (``highway tag'') for most roads.
We determined an average road width for each category on a small
subset of the data, and validated it on a larger subset (manually,
one-off).
This simple strategy works reasonably well, with a mean error of
$\approx\,$11 pixels for the road boundary, compared to $\approx\,$100
pixels of road width\footnote{Average deviation based on 10 random
  samples of Potsdam, Chicago, Paris, and Zurich.}.
In (very rare) cases where the ad-hoc procedure produced label
collisions, pixels claimed by both building and road were assigned to
buildings. Pixels neither labeled building nor road form the
background class. Examples of images overlaid with automatically
generated OSM labels are shown in Fig.\ref{fig:img_ground_truth}.

\subsection{Neural network architecture}

We use a variant of fully convolutional networks (FCN) in this paper,
see Fig.~\ref{fig:fcn_visualization}.
Following the standard neural network concept, transformations are
ordered in sequential layers that gradually transform the pixel values
to label probabilities. Most layers implement learned convolution
filters, where each neuron at level $l$ takes its input values only
from a fixed-size, spatially localized window $\mathcal{W}$ in the
previous layer $(l-1)$, and outputs a vector of differently weighted
sums of those values, $c^l=\sum_{i\in\mathcal{W}}w_i
c^{l-1}_i$. Weights $w_i$ are shared across all neurons of a layer,
which reflects the shift-invariance of the image signal and
drastically reduces the number of parameters. Each convolutional layer
is followed by a rectified linear unit ($ReLU$)
$c^l_\text{rec}=\max(0,c^l)$, which simply truncates all negative
values to $0$ and leaves positive values unchanged
\cite{nair2010rectified}\footnote{Other non-linearities are sometimes
  used, but $ReLU$ has been shown to facilitate training
  (backpropagation) and has become the de-facto standard.}.
Convolutional layers are interspersed with Max-Pooling layers that
downsample the image and retain only the maximum value inside a
(2$\times$2) neighborhood. The downsampling increases the receptive
field of subsequent convolutions, and lets the network learn
correlations over a larger spatial context. Moreover, max-pooling
achieves local translation invariance at object level.
The outputs of the last convolutional layers (which are very big to
capture global context, equivalent to a fully connected layer of
standard CNNs) is converted to a vector of scores for the three
target classes.
These score maps are of low resolution, hence they are gradually
upsampled again with convolutional layers using a stride of only
$\frac{1}{2}$ pixel.%
\footnote{This operation is done by layers that are usually called ``deconvolution layers'' in the literature~\citep{long2015} (and also in Fig.~\ref{fig:fcn-4s-1_architecture}) although the use of this terminology has been critized since most implementations do not perform a real deconvolution but rather a transposed convolution.} %
Repeated downsampling causes a loss of high-frequency content, which
leads to blurry boundaries that are undesirable for pixel-wise
semantic segmentation.
To counter this effect, feature maps at intermediate layers are merged
back in during upsampling (so-called ``skip connections'', see
Fig.~\ref{fig:fcn_visualization}).
The final, full-resolution score maps are then converted to label
probabilities with the $\mathop{softmax}$ function.

\subsection{Implementation Details}
\begin{figure*}
	\centering
	\includegraphics[width=0.8\linewidth]{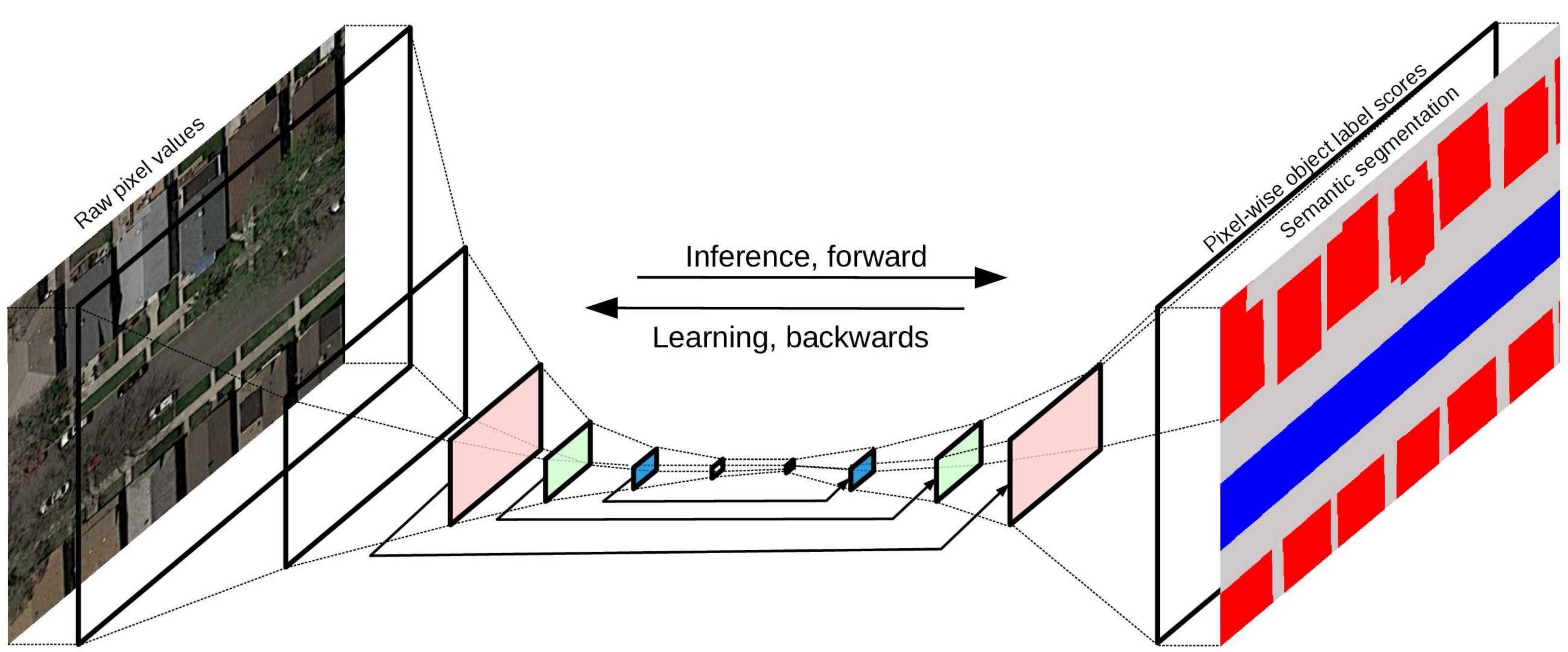}
	\caption{Conceptual illustration of the data flow through our variant of a fully convolutional network (FCN), which is used for the
          semantic segmentation of aerial images. Three
          skip-connections are highlighted by pale red, pale green,
          and pale blue, respectively. Note that we added a third (pale red) skip connection in addition to the original ones (pale green, pale blue) of~\cite{long2015}.}
	\label{fig:fcn_visualization}
\end{figure*}

\begin{figure*}
	\centering
	\includegraphics[width=0.75\linewidth]{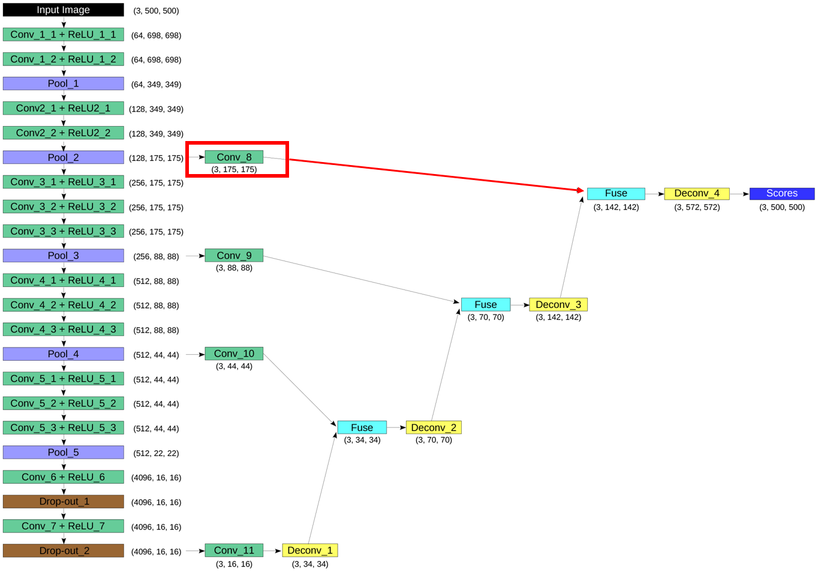}
	\caption{Our FCN architecture, which adds one more skip-connection (after Pool\_2, shown red) to the original model of \cite{long2015}. Neurons form a three-dimensional structure per layer: dimensions are written in brackets, where the first number indicates the amount of feature channels, second and third represent spatial dimensions.}
	\label{fig:fcn-4s-1_architecture}
\end{figure*}

The FCN we use is an adaptation of the architecture proposed
in~\cite{long2015}, which itself is largely based on the VGG-16
network architecture~\cite{simonyan2015very}.
In our implementation, we slightly modify the original FCN and
introduce a third skip connection (marked red in
Fig.~\ref{fig:fcn_visualization}), to preserve even finer image
details.
We found that the original architecture, which has two
skip-connections after Pool\_3 and Pool\_4
(cf.\ Fig.~\ref{fig:fcn-4s-1_architecture}) was still not delivering
sufficiently sharp edges.
The additional, higher-resolution skip connection consistently
improved the results for our data, see Sec.~\ref{sec:results}.
Note that adding the third skip-connection does not increase the
total number of parameters but, on the contrary, slightly reduces it
(\cite{long2015}: $134'277'737$, ours: $134'276'540$; the small
difference is due to the decomposition of the final upsampling kernel
into two smaller ones).


\subsection{Training}

All model parameters are learned by minimising a multinomial logistic
loss, summed over the entire 500$\times$500 pixel patch that
serves as input to the FCN.
Prior to training/inference, intensity distributions are centred
independently per patch by subtracting the mean, separately for each
channel (RGB).

All models are trained with stochastic gradient descent with a
momentum of 0.9, and minibatch size of 1 image. Learning rates always
start from 5$\times$ 10$^{-9}$ and are reduced by a factor of 10 twice
when the loss and average $F_1$ scores stopped improving. The learning
rates for biases of convolutional layers were doubled with respect to
learning rates of the filter weights. Weight decay was set to 5
$\times$ 10$^{-4}$, dropout probability for neurons in layers ReLU\_6
and ReLU\_7 was always 0.5.

Training was run until the average \(F_1\)-score on the validation
dataset stopped improving, which took between $45000$ and $140000$
iterations (3.5-6.5 epochs).
Weights were initialized as in Glorot et
al.~\cite{glorot2010understanding}, except for experiments with
pre-trained weights.
It is a common practice in deep learning to publish pre-trained models
together with source code and paper, to ease repeatability of results
and to help others avoid training from scratch. Starting from
pre-trained models, even if these have been trained on a completely
different image dataset, often improves performance, because low-level
features like contrast edges and blobs learned in early network layers
are very similar across different kinds of images.

We will use two different forms of pre-training. Either we rely on
weights previously learned on the Pascal VOC
benchmark~\citep{everingham2010pascal} (made available by the authors
of~\cite{long2015}). Or we pre-train ourselves with OSM data. In the
experiments section it is always specified whether we use VOC, OSM or
no pre-training at all.

\section{EXPERIMENTS}\label{sec:EXPERIMENTS}

We present extensive experiments on four large data sets of
different cities to explore the following scenarios:
\begin{itemize}
\item
\textit{Complete substitution:} Can semantic segmentation be learned
without any manual labeling? What performance can be achieved using
only noisy labels gleaned from OSM?
\item
  \textit{Augmentation:} Will pre-training with large-scale OSM data
  and publicly available images improve the segmentation of a
  project-specific data set of independently acquired images and
  labels?
	\item
 \textit{Partial substitution:} Can pre-training with large-scale OSM
  labels replace a substantial part of the manual labeling effort?
  Phrased differently, can a generic model learned from OSM be adapted
  to a specific location and data source with only little dedicated
  training data?
\end{itemize}

We provide a summary of the results and explicit answers to these questions at the very end of this section. 
Note that all experiments are designed to investigate different aspects of the hypotheses made in the introduction. We briefly remind and thoroughly validate all hypotheses based on results of our experiments in the conclusion.
  
\subsection{Datasets}\label{sec:DATASETS}

\begin{figure}
	\centering
	\includegraphics[width=0.99\linewidth]{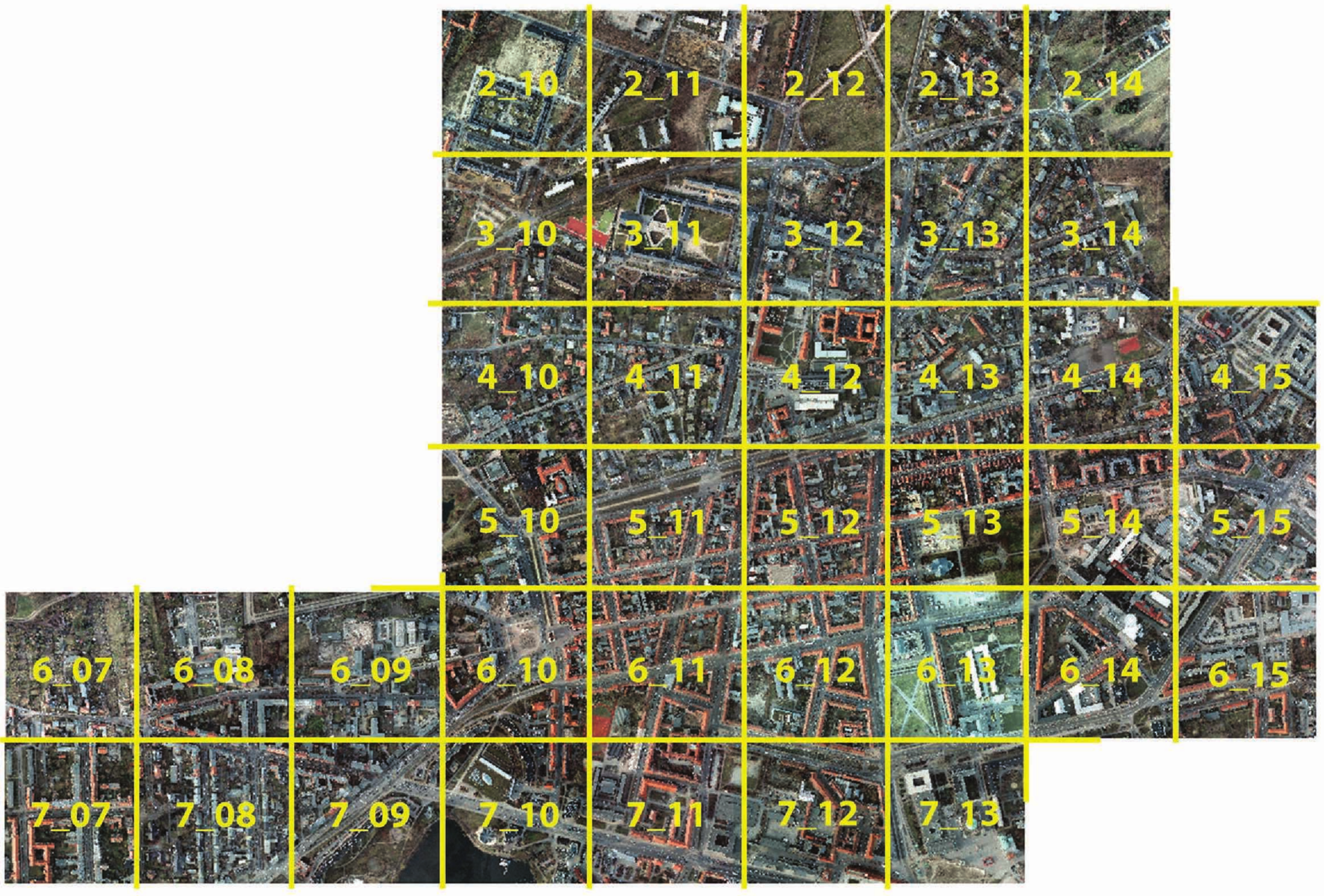}\\
	\caption{Overview of the ISPRS Potsdam dataset. The aerial
          images shown are those provided by the ISPRS
          benchmark~\cite{rottensteiner2013}.}
  \label{fig:potsdam_dataset}
\end{figure}

Four large datasets were downloaded from Google Maps and OSM, for
the cities of Chicago, Paris, Zurich, and Berlin.
Additionally we also downloaded a somewhat smaller dataset for the city
of Potsdam. For this location, a separate image set and high-accuracy
ground truth is available from the ISPRS semantic labeling
benchmark~\cite{rottensteiner2013}.
Table~\ref{tab:ground_truth_data} specifies the coverage (surface
area), number of pixels, and ground sampling distance of each dataset.
Example images and segmentation maps of Paris and Zurich are shown in
Figure~\ref{fig:img_ground_truth}. In Fig~\ref{fig:potsdam_dataset} we
show the full extent of the Potsdam scene, dictated by the available
images and ground truth in the ISPRS benchmark.
OSM maps and aerial images from Google Maps where downloaded and cut
to cover exactly the same region to ensure a meaningful comparison --
this meant, however, that the dataset is an order of magnitude smaller
than what we call ``large-scale'' for the other cities.
The ISPRS dataset includes a portion (images $x\_13, x\_14, x\_15$ on
the right side of Fig.~\ref{fig:potsdam_dataset}), for which the
ground truth is withheld to serve as test set for benchmark
submissions.
We thus use images $2\_12$, $6\_10$, and $7\_11$ as test set, and the
remaining ones for training.
The three test images were selected to cover different levels of
urban density and architectural layout. This train-test split
corresponds to $1.89 km^2$ of training data, respectively $0.27 km^2$
of test data.

The ISPRS semantic labeling challenge aims at land-cover
classification, whereas OSM represents land-use. In particular, the
benchmark ground truth does not have a label \textit{street}, but
instead uses a broader class \textit{impervious surfaces}, also
comprising sidewalks, tarmacked courtyards etc. Furthermore, it labels
overhanging tree canopies that occlude parts of the impervious ground
(including streets) as \textit{tree}, whereas \textit{streets} in the OSM
labels include pixels under trees.
Moreover, images in the ISPRS benchmark are ``true'' orthophotos
rectified with a DSM that includes buildings, whereas Google images
are conventional orthophotos. corrected only for terrain-induced
distortions with a DTM. Building facades remain visible and roofs
are shifted from the true footprint.
To facilitate a meaningful comparison, we have manually re-labeled the
ISPRS ground truth to our target categories \textit{street}, and
\textit{background}, matching the land-use definitions extracted from
OSM. The category \textit{building} of the benchmark  ground truth
remains unchanged.
To allow for a direct and fair comparison, we down-sample the ISPRS Potsdam 
data, which comes at a ground sampling distance (GSD) of $5~cm$, to the 
same GSD as the Potsdam-Google data ($9.1~cm$). 
\begin{table}
	\small
	\begin{center}
    		\begin{tabular}{| l | c | c | c |}
	    		\hline
    			& Coverage & No. of pixels & GSD \\ \hline
			Chicago & $50.6~km^2$ & \(4.1 \times 10^9\) & $11.1~cm$ \\
			Paris & $60.3~km^2$ & \(6.3 \times 10^9\) & $9.8~cm$ \\ 
			Zurich & $36.2~km^2$ & \(3.5 \times 10^9\) &  $10.1~cm$ \\ 
			Berlin & $10.6~km^2$ & \(1.28 \times 10^9\) &  $9.1~cm$ \\ \hline
			Potsdam-Google & $2.16~km^2$ & \(2.60 \times 10^8\) & $9.1~cm$ \\
			Potsdam-ISPRS & $2.16~km^2$ & \(2.60 \times 10^8\) & $9.1~cm$ \\ \hline
		\end{tabular}
	\end{center}
	\caption{Statistics of the datasets used in the
          experiments. Note that we down-sampled the original
          Potsdam-ISPRS ($GSD=5~cm$) to the resolution of the
          Potsdam-Google data ($GSD=9.1~cm$) for all experiments.}
	\label{tab:ground_truth_data}
\end{table}

For all datasets, we cut the aerial images as well as the
corresponding label maps into non-overlapping tiles of size 500
$\times$ 500 pixels. The size was determined in preliminary
experiments, to include sufficient geographical context while keeping
FCN training and prediction efficient on a normal single-GPU desktop
machine.
Each dataset is split into mutually exclusive training, validation and
test regions.
During training, we monitor the loss (objective function) not only on the
training set, but also on the validation set to prevent overfitting.%
\footnote{This is standard practice when training deep neural networks.}


\subsection{Results and discussion}\label{sec:results}

First, we validate our modifications of the FCN architecture, by
comparing it to the original model of~\cite{long2015}.
As error metrics, we always compute precision, recall and $F_1$-score,
per class as well as averaged over all three classes.
Precision is defined as the fraction of predicted labels that are 
correct with respect to ground truth, recall is the fraction of true labels that are correctly
predicted.
The $F_1$-score is the harmonic mean between precision and recall. It
combines the two competing goals into a scalar metric and is widely
used to assess semantic segmentation. It also serves as our primary
error measure.
\begin{figure*}
	\centering
        \begin{tabular}{@{}c@{\hspace{5mm}}c@{\hspace{5mm}}}
	\includegraphics[width=0.42\linewidth]{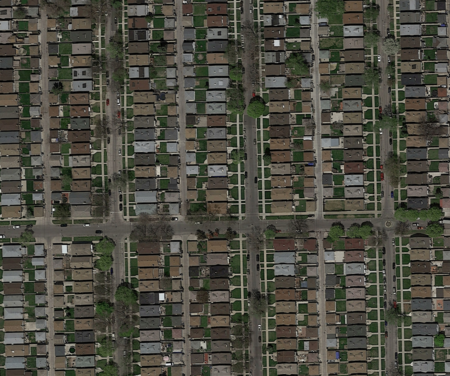} & 
	\includegraphics[width=0.42\linewidth]{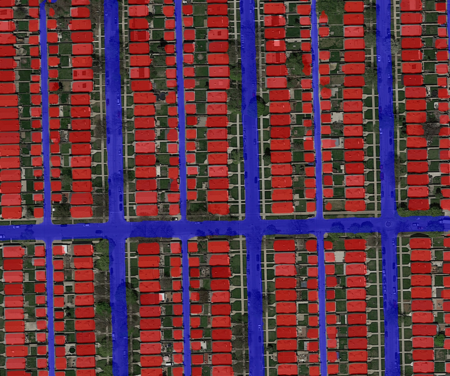} \\
	(a) & (b)\\
	    \end{tabular}
	\caption{FCN trained on Google Maps imagery and OSM labels of
          Chicago: (a) Original aerial image and (b) overlaid with
          classification result.}
  \label{fig:results_chicago}
\end{figure*}
Quantitative results are shown in
Table~\ref{tab:fcn_performance_f1}, an example result for Chicago is 
shown in Figure~\ref{fig:results_chicago}. Our architecture with the
additional early skip-connection outperforms its counterpart slightly 
but consistently on average, albeit only by $\approx 1$ percent 
point. 
Note that this performance improvement also comes with the benefit of lower run-times. Our architecture consistently needs $\geq 30\%$ less time 
for training compared to the original architecture of~\citep{long2015} (see Table~\ref{tab:fcn_performance_f1}).
%

\begin{table*}
	\small
	\begin{center}
    		\begin{tabular}{| l | c | c | c | c | c | c |}
	    		\hline
	    		& \multicolumn{2}{c|}{Chicago} & \multicolumn{2}{c|}{Paris} & \multicolumn{2}{c|}{Zurich} \\ \hline
	    		               & ~\cite{long2015} (15.7h) & Ours (10.5h)  & ~\cite{long2015} (18.3h) & Ours (7.6h) & ~\cite{long2015} (15.5h) & Ours (6.2h)\\ \hline 
			\(F_1\) average    & 0.840                    & {\bf0.855}    & 0.774                    & \textbf{0.776}               & 0.804                    & \textbf{0.810} \\ \hline 
			\(F_1\) building   & 0.823                    & {\bf0.837}    & 0.821                    & \textbf{0.822}               & \textbf{0.824}           & 0.823 \\ \hline 
			\(F_1\) road       & 0.821                    & {\bf0.843}    & 0.741                    & \textbf{0.746}               & 0.695                    & 0.707 \\ \hline 
			\(F_1\) background & 0.849                    & {\bf0.861}    & \textbf{0.754}           & \textbf{0.754}               & \textbf{0.894}           & 0.891 \\ \hline 
		\end{tabular}
	\end{center}
	\caption{Comparison between our adapted FCN and the original architecture of~\cite{long2015}, for three large city datasets. Numbers in brackets indicate training times for the original FCN architecture of~\citep{long2015} and ours for all data sets if trained from scratch without any pre-training to facilitate a fair comparison (on a standard, stand-alone PC with i7 CPU, 2.7 GHz, 64 GB RAM and NVIDIA TITAN-X GPU with 12 GB RAM).}
	\label{tab:fcn_performance_f1}
\end{table*}
\begin{figure*}
	\centering
        \begin{tabular}{@{}c@{\hspace{5mm}}c@{\hspace{5mm}}c@{\hspace{5mm}}c@{\hspace{5mm}}}
	\includegraphics[width=0.22\linewidth]{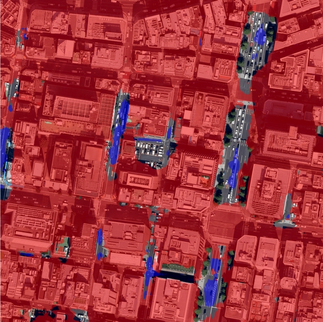} & 
	\includegraphics[width=0.22\linewidth]{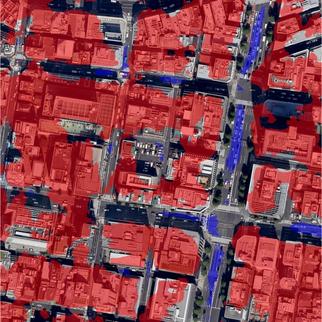} &
	\includegraphics[width=0.22\linewidth]{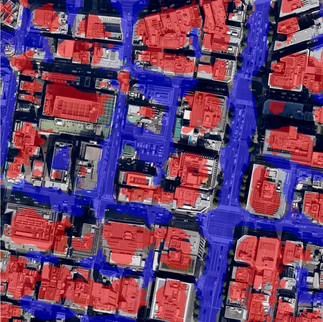} &
	\includegraphics[width=0.22\linewidth]{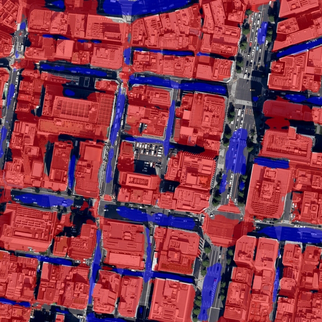} \\
	(a) & (b) & (c) & (d)\\
	    \end{tabular}
	\caption{Classification results and average $F^{1}$-scores of the Tokyo scene with a model trained on
          (a) Chicago ($F^{1}:$ 0.485), (b) Paris ($F^{1}:$ 0.521), (c) Zurich ($F^{1}:$ 0.581), (d) all three ($F^{1}:$ 0.644).}
  \label{fig:results_tokyo}
\end{figure*} 
%
\begin{table*}
	\small
	\begin{center}
    		\begin{tabular}{| l | l |c | c | c |c | c |}
	    		\cline{3-7}
	    		\multicolumn{2}{c|}{}							  &\# ISPRS images & pre-training& OSM       & Google images& test data \\ \hline 
{\bf Ia}  & ISPRS baseline        		        &3        	     & no          &-          &-             & ISPRS     \\ \hline
{\bf Ib}	& ISPRS baseline pre-trained	      &3               & yes         &-          &-             & ISPRS     \\ \hline
{\bf II}	& ISPRS gold standard   			      &21              & yes \& no   &-          &-             & ISPRS     \\ \hline 
{\bf IIIa}& Google/OSM baseline Potsdam  			&-               & no          &P          &P             & OSM+Google\\ \hline
{\bf IIIb}& Google/OSM baseline Potsdam+Berlin&-               & yes         &P, B       &P, B          & OSM+Google\\ \hline\hline
{\bf IV}  & Complete substitution 			      &21              & no          &P          &-             & ISPRS     \\ \hline
{\bf V}   & Augmentation          			      &21              & yes         &B, Z, C, P &B, Z, C, P    & ISPRS     \\ \hline 
{\bf VI}  & Partial substitution  			      &3               & yes         &B, Z, C, P &B, Z, C, P    & ISPRS     \\ \hline 
                \end{tabular}
        \end{center}
        \caption{Overview of different experimental setups we use to validate our hypothesis made in the introduction. We abbreviate Berlin (B), Zurich (Z), Chicago (C), and Potsdam (P). All entries refer to the training setup except the most right column, which indicates data used for testing. Quantitative results for all experiments are given in Table~\ref{tab:potsdam_results}.} 
	\label{tab:exp_overview}
\end{table*}
Another interesting finding is in terms of transfer learning, 
in the sense that training a model over multiple
cities, with both different global scene structure and different
object appearance, can help better predict a new, previously unseen
city. 
This again emphasizes the improved generalization ability that benefits 
from the increased amount of weak labels, in contrast to traditional supervised 
approaches with smaller label sets.
We train the FCN on Zurich, Paris, and Chicago and predict
Tokyo. We compare the results with those from training on only a
single city (Fig~\ref{fig:results_tokyo}).
It turns out that training over multiple, different cities helps the
model to find a more general, ``mean'' representation of what a city
looks like. Generalising from a single city to Tokyo clearly performs
worse (Fig.~\ref{fig:results_tokyo}(a,b,c)) than generalising from
several different ones (Fig.~\ref{fig:results_tokyo}(d)).
This indicates that FCNs are indeed able to learn location-specific
urbanistic and architectural patterns; but also that supervision with
a sufficiently diverse training set mitigates this effect and still
lets the system learn more global, generic patterns that support
semantic segmentation in different geographic regions not seen at all
during training.

For experiments on the ISPRS Potsdam data set, we first compute three
baselines. 
For an overview of the setup of all experiments described 
in the following, please refer to Table~\ref{tab:exp_overview} whereas 
quantitative results are given in Table~\ref{tab:potsdam_results}. 
\begin{figure*}
	\centering
        \begin{tabular}{@{}c@{\hspace{5mm}}c@{\hspace{5mm}}c@{\hspace{5mm}}}
	\includegraphics[width=0.25\linewidth]{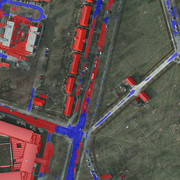} & 
	\includegraphics[width=0.25\linewidth]{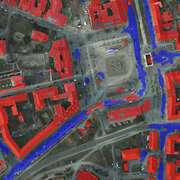} &
	\includegraphics[width=0.25\linewidth]{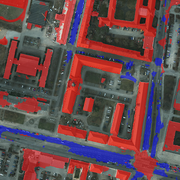} \\
	(a) & (b) & (c)\\
	\includegraphics[width=0.25\linewidth]{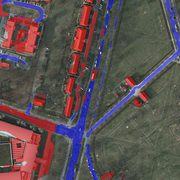} & 
	\includegraphics[width=0.25\linewidth]{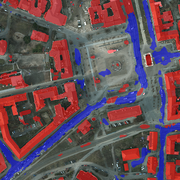} &
	\includegraphics[width=0.25\linewidth]{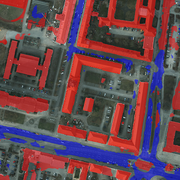} \\
	(d) & (e) & (f)\\
	\includegraphics[width=0.25\linewidth]{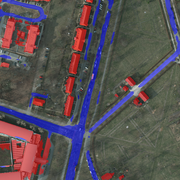} & 
	\includegraphics[width=0.25\linewidth]{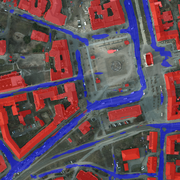} &
	\includegraphics[width=0.25\linewidth]{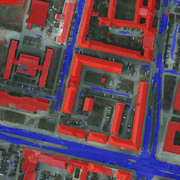} \\
	(g) & (h) & (i)\\
	    \end{tabular}
	\caption{Baseline experiments, (a,b,c):
            Baseline {\bf Ia} trained on three ISPRS
            images without pre-training. (d,e,f): Baseline
            {\bf Ib} trained on three ISPRS images with
            pre-training on Pascal VOC. (g,h,i): Gold standard
            {\bf II} trained on 21 ISPRS images.}
  \label{fig:results_baselines_1}
\end{figure*}

\paragraph*{({\bf\itshape I}) Baseline with ISPRS data}
First, we follow the conventional semantic segmentation baseline and
apply our FCN model to the ISPRS benchmark to establish a baseline
with conventional, hand-labeled ground truth.
As a training set of realistic size we use three completely labelled
images from the ISPRS Potsdam benchmark (3.25$\cdot$10$^7$ pixels /
27$\,$ha).
This setup {\bf Ia} achieves $0.764$ average
\(F_1\)-score over the three classes if we train our FCN from scratch,
i.e., weights initialized randomly as
in~\cite{glorot2010understanding}
(Fig.~\ref{fig:results_baselines_1}(a,b,c)).
A widely used practice is to start from a pre-trained model that has
been learned from a very large dataset, especially if the dedicated
training data is limited in size. We thus compute baseline
{\bf Ib}, where we start from a model trained on the
Pascal VOC benchmark and fine-tune on the three ISPRS Potsdam
images. As expected this boosts performance, to $0.809$ average
\(F_1\)-score (Fig.~\ref{fig:results_baselines_1}(d,e,f)).

\paragraph*{({\bf\itshape II}) Gold standard with ISPRS data}
Second, we repeat the same experiment, but use all of the available
training data, i.e., we train on all 21 available training images
(2.275$\cdot$10$^8$ pixels / 189$\,$ha).
This setup serves as a ``gold standard'' for what is achievable with
the conventional pipeline, given an unusually large amount of costly
high-quality training labels.
It simulates a project with the luxury of $>$200 million hand-labelled
training pixels over a medium-sized city (which will rarely be the
case in practice).
It achieves an $F_1$-score of $0.874$ if trained from scratch
(Fig.~\ref{fig:results_baselines_1}).  The significant improvement of
11, respectively 6 percent points shows that our ``standard''
baselines {\bf Ia} and {\bf Ib} are
still data-limited, and can potentially be improved significantly with
additional training data.
As a sanity check, we also ran the same experiment with all 21 ISPRS
images \textit{and} pre-training from Pascal VOC. This only marginally
increases the average $F_1$-score to $0.879$.

We note that baseline {\bf II} is not directly
comparable with the existing benchmark entries, since we work with a
reduced class nomenclature and modified ground truth, and do not
evaluate on the undisclosed test set.
But it lies in a plausible range, on par with or slightly below the
\textit{impervious ground} and \textit{building} results of the
competitors, who, unlike us, also use the DSM.

\begin{figure*}
	\centering
        \begin{tabular}{@{}c@{\hspace{5mm}}c@{\hspace{5mm}}c@{\hspace{5mm}}}
	\includegraphics[width=0.25\linewidth]{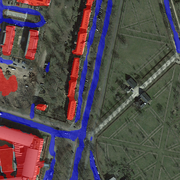} & 
	\includegraphics[width=0.25\linewidth]{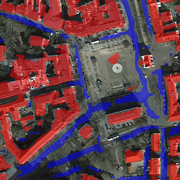} &
	\includegraphics[width=0.25\linewidth]{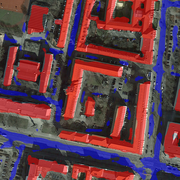} \\	
	(a) & (b) & (c)\\
	\includegraphics[width=0.25\linewidth]{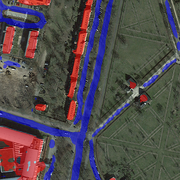} & 
	\includegraphics[width=0.25\linewidth]{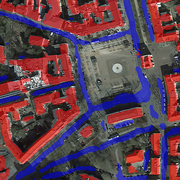} &
	\includegraphics[width=0.25\linewidth]{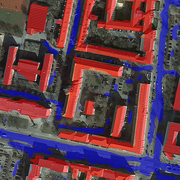} \\	
	(d) & (e) & (f)\\
	    \end{tabular}
	\caption{Baseline experiments, (a,b,c): Baseline
          {\bf IIIa} with Google Maps images and OSM
          Maps from only Potsdam. (d,e,f): Baseline {\bf
            IIIb} with Google Maps images and OSM Maps and training on
          Potsdam and Berlin.}
  \label{fig:results_baselines_2}
\end{figure*}

\paragraph*{({\bf\itshape III}) Baseline with Google Maps images and OSM Maps}

The next baseline {\bf IIIa} trains on Google aerial
images using OSM map data as ground truth. The same 189$\,$ha as in
baseline {\bf II} are used for training, and the model achieves an
$F_1$-score of $0.777$ if tested on Google aerial images and OSM
ground truth (Fig.~\ref{fig:results_baselines_2}(a,b,c)).
This baseline has been added as a sanity check to verify that the
previously observed potential of the open data sources is confirmed
also for Potsdam.
We point out that the experiment is somewhat problematic and not
comparable to baseline {\bf II}, in that it inevitably
confounds several effects: the drop in performance may in part be due
to the larger amount of noise in the training labels; but further
possible reasons include on the one hand the inferior image quality of
the Google Maps images, c.f.\ cast shadows and ortho-rectification
artifacts in Fig.~\ref{fig:results_baselines_2}(b,c); and on the other
hand the noise in the OSM-based test
labels.\footnote{We also test the same model on Google
    aerial images with ISPRS labels, which leads to a slight
    performance drop to $0.759$.  This is not surprising,
    because labels have been acquired based on the ISPRS images and do
    not fit as accurately to the Google images.} %
Recall that the same setup achieved $0.810$ for the architecturally
comparable Zurich, and $0.827$ for the more schematic layout of
Chicago.
This suggests that a part of the drop may be attributed to the smaller
training set, respectively that noisy OSM labels should be used in
large quantities.
To verify this assumption we repeat the experiment, but greatly extend
the training dataset by adding the larger city of Berlin, which is
immediately adjacent to Potsdam. This baseline {\bf
  IIIb} increases performance by $2$ percent points to $0.797$
(Fig.~\ref{fig:results_baselines_2}(d,e,f)), which is only slightly
below performance on Zurich ($0.810$). It shows that training data
size is a crucial factor, and that indeed city-scale (though noisy)
training data helps to learn better models.

%
Qualitatively, one can see that the model trained on OSM has a
tendency to miss bits of the road, and produces slightly less accurate
and blurrier building outlines.
\begin{figure*}
	\centering
        \begin{tabular}{@{}c@{\hspace{5mm}}c@{\hspace{5mm}}c@{\hspace{5mm}}}
	\includegraphics[width=0.25\linewidth]{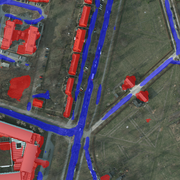} & 
	\includegraphics[width=0.25\linewidth]{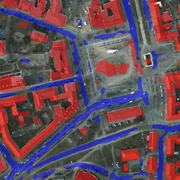} &
	\includegraphics[width=0.25\linewidth]{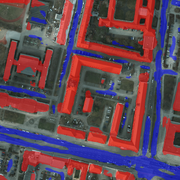} \\
	(a) & (b) & (c)\\
	\includegraphics[width=0.25\linewidth]{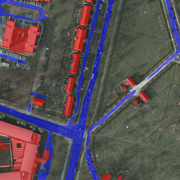} & 
	\includegraphics[width=0.25\linewidth]{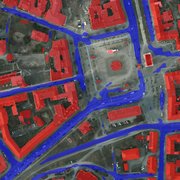} &
	\includegraphics[width=0.25\linewidth]{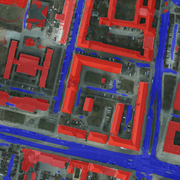} \\
	(d) & (e) & (f)\\
	\includegraphics[width=0.25\linewidth]{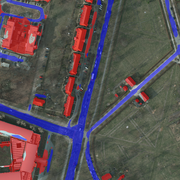} & 
	\includegraphics[width=0.25\linewidth]{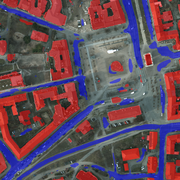} &
	\includegraphics[width=0.25\linewidth]{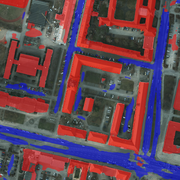} \\
	(g) & (h) & (i)\\
	    \end{tabular}
	\caption{(a,b,c): Complete substitution
          ({\bf IV}) of manual labels, train from
          scratch on ISPRS images and OSM labels of Potsdam (no
          pre-training). (d,e,f): Augmentation
          ({\bf V}) with open data, pre-train on
          Chicago, Paris, Zurich, and Berlin and re-train on all 21
          ISPRS training images with pixel-accurate ground
          truth. (g,h,i): Partial substitution
          ({\bf VI}) of manual labels, pre-train on
          Chicago, Paris, Zurich, and Berlin and re-train on 3 ISPRS
          images with pixel-accurate ground truth.}
  \label{fig:results_pretraining}
\end{figure*}

\begin{figure*}
	\centering
        \begin{tabular}{@{}c@{\hspace{5mm}}c@{\hspace{5mm}}c@{\hspace{5mm}}}
	\includegraphics[width=0.25\linewidth]{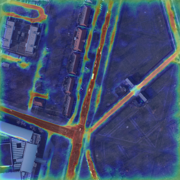} & 
	\includegraphics[width=0.25\linewidth]{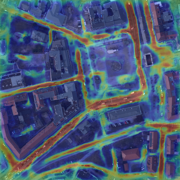} &
	\includegraphics[width=0.25\linewidth]{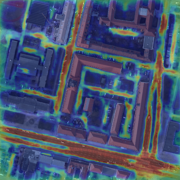} \\	
	(a) & (b) & (c)\\
	\includegraphics[width=0.25\linewidth]{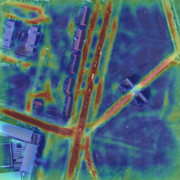} & 
	\includegraphics[width=0.25\linewidth]{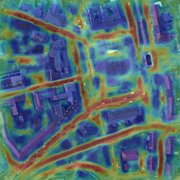} &
	\includegraphics[width=0.25\linewidth]{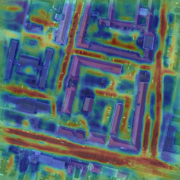} \\	
	(d) & (e) & (f)\\
	    \end{tabular}
	\caption{Probability maps for road extraction of the gold
          standard baseline {\bf II} (a,b,c); and
          complete substitution {\bf IV} without any
          manual labels (d,e,f). Road probabilities range from red
          (high) to blue (low).}
  \label{fig:scoremaps_ii_iv}
\end{figure*}

\paragraph*{({\bf\itshape IV}) Complete substitution of manual labels}
Next, we evaluate the extreme setting where we do not have any
high-accuracy labels and completely rely on OSM as source of training
data.
We thus train our FCN on the ISPRS Potsdam images, but use OSM map
data as ground truth. The predictions for the ISPRS test images are
then evaluated with the manual high-accuracy ground truth from the
benchmark.
In other words, this experiments quantifies how accurate predictions
we can expect if training from OSM labels for a limited,
project-specific image set: since the ISPRS dataset does not provide
more images, one cannot augment the training set further, even though
a lot more OSM data would be available.
This set up achieves an $F_1$-score of $0.779$, beating baseline {\bf
  Ia} by 1.5 percent points.  We conclude that \textit{larger amounts
  of noisy, automatically gleaned training data can indeed completely
  replace small amounts of highly accurate training data}, saving the
  associated effort and cost.
The result however does stay 3 percent points behind baseline {\bf
  Ib}, which shows that even all of Potsdam is not large enough to
replace pre-training with large-scale data, which will be addressed in
experiment {\bf VI}.
Compared to baseline {\bf II}, i.e.\ training with
equally large quantities of pixel-accurate labels, performance drops
by 10 percent points.
The visual comparison between baseline {\bf II} in
Fig.~\ref{fig:results_baselines_1}(g,h,i) and {\bf IV}
in Fig.~\ref{fig:results_pretraining}(a,b,c) shows that buildings are
segmented equally well, but roads deteriorate significantly. This is
confirmed by the $F_1$-scores in Table~\ref{tab:potsdam_results}.
An explanation is the noise in the guessed road width (as also pointed out by~\citep{maggiori2016}) in the training
data ($\approx 23$ pixels on average, for an average road width of
$\approx 100$ pixels). It leads to washed-out evidence near the road
boundaries, which in turn weakens the overall evidence in the
case of narrow or weakly supported roads.
This effect can be observed visually by comparing probability maps of
{\bf II} and {\bf IV} in
Fig.~\ref{fig:scoremaps_ii_iv}. Road probabilities appear much sharper
at road edges for baseline {\bf II} trained with
pixel-accurate ISPRS groundtruth
(Fig.~\ref{fig:scoremaps_ii_iv}(a,b,c)) compared to {\bf
  IV} trained with noisy OSM ground truth
(Fig.~\ref{fig:scoremaps_ii_iv}(d,e,f)).

\paragraph*{({\bf\itshape V}) Augmentation with open data}

With experiment {\bf V} we aim to assess whether
pre-training from even larger amounts of open data from other sites
can further improve the gold-standard {\bf II}, by
providing a sort of ``generic background'' for the problem, in the
spirit of pre-trained computer vision models such as
VGG~\cite{simonyan2015very} or Alexnet~\cite{krizhevsky2012imagenet}.
We first train the FCN model on Google/OSM data of Chicago, Paris,
Zurich, and Berlin, and use the resulting network weights as initial
value, from which the model is tuned for the ISPRS data, using all the
21 training images as in baseline {\bf II}.
The pre-training boosts performance, albeit only by 1 percent
point. Even if one has a comfortable amount of accurate training data
at hand, it appears potentially useful to pre-train with freely
available data. In future work it may be useful to experiment with even larger amounts of open data.

A visual comparison of Fig.~\ref{fig:results_baselines_1}(g,h,i) and
Fig.~\ref{fig:results_pretraining}(d,e,f) shows small improvements for
both the roads and the buildings, in all three tiles.
This effect shows up quantitatively with an improvement in
\(F_1\)-score of the \textit{road} class, which reaches $0.825$, up
from $0.764$ in baseline {\bf II}. On the other hand,
buildings are detected equally well, no further improvement can be
noticed. A possible interpretation is that complex network structures
with long-range dependencies are hard to learn for the classifier, and
thus more training data helps. Locally well-defined, compact objects
of similar shape and appearance are easier to learn, so further
training data does not add relevant information.
%
\begin{table*}
	\small
	\begin{center}
    		\begin{tabular}{| l | l |c | c | c ||c | c | c | c |}
	    		\cline{3-9}
	    		\multicolumn{2}{c|}{}							  &av.~\(F_1\)      &av.~Precision    &av.~Recall       &\(F_1\) Building &\(F_1\) Road     &\(F_1\) Background & train time [h] \\ \hline 
{\bf Ia}  & ISPRS baseline        		        &0.764            &0.835            &0.704            &0.793            &0.499            &0.883              & 16 \\ \hline
{\bf Ib}	& ISPRS baseline pre-trained	      &0.809            &0.853            &0.770            &0.830            &0.636            &\textit{0.904} 		& 16 \\ \hline
{\bf II}	& ISPRS gold standard   			      &\underline{0.874}&\textbf{0.910}   &\underline{0.841}&\textbf{0.913}   &\underline{0.764}&\textbf{0.923} 		& 16 \\ \hline 
{\bf IIIa}& Google/OSM baseline P  						&0.777            &0.799            &0.756            &0.832            &0.631            &0.845  						& 16 \\ \hline
{\bf IIIb}& Google/OSM baseline P+B						&0.797            &0.819            &0.776            &0.828            &0.698            &0.858  						& 32 \\ \hline\hline
{\bf IV}  & Complete substitution 			      &0.779            &0.801            &0.758            &0.796            &0.667            &0.860  						& 16 \\ \hline
{\bf V}   & Augmentation          			      &\textbf{0.884}   &\underline{0.898}&\textbf{0.870}   &\underline{0.900}&\textbf{0.825}   &\underline{0.922}	& 78 \\ \hline 
{\bf VI}  & Partial substitution  			      &\textit{0.837}   &\textit{0.860}   &\textit{0.816}   &\textit{0.863}   &\textit{0.736}   &0.899  						& 78 \\ \hline 
                \end{tabular}
        \end{center}
        \caption{Results of experiments with the Potsdam data set. The three left columns are average values over all classes, the right three columns give per class $F_1$-scores. \textbf{Best results} across all variants are written in bold font, \underline{second best results} are underlined, and \textit{third best results} have italic fond type. All experiments (and run-times) were computed on a standard, stand-alone PC with i7 CPU, 2.7 GHz, 64 GB RAM and NVIDIA TITAN-X GPU with 12 GB RAM. Like in Tab.~\ref{tab:exp_overview}, P is short for Potsdam whereas B is short for Berlin.} 
	\label{tab:potsdam_results}
\end{table*}

\paragraph*{({\bf\itshape VI}) Partial substitution of manual labels}

The success of pre-training in previous experiments raises the question - also asked in~\citep{maggiori2017} - of whether one could reduce the annotation effort and use a smaller hand-labelled training set, in conjunction with large-scale OSM labels.
An alternative view is as a domain adaptation problem, where the
classifier is trained on Google Maps images, and then re-targeted to
ISPRS images with only few training samples.
The hope is that the large amount of OSM training data would already
allow the classifier to learn basic aerial image statistics and urban
scene structures.
Then, only a small additional training set would suffice to adapt it
to the different spectral properties.
In experiment {\bf VI} we therefore first train the FCN
on the combined Google / OSM data of Chicago, Paris, Zurich, and
Berlin. This part is the same as in experiment {\bf
  V}. Then, we use only the small set of training images and labels
from baseline {\bf I} to tune it to the ISPRS images of
Potsdam.
Performance increases by 7 percent points to $0.837$ over baseline
{\bf Ia}, where the model is trained from scratch on the
same high-accuracy labels.
We conclude that if only a limited quantity of high-quality training
data is available, pre-training on free data brings even larger
relative benefits, and can be recommended as general practice, which is in line with the findings reported in~\citep{maggiori2017}.

Importantly, experiment {\bf VI} also outperforms
baseline {\bf Ib} by almost 3 percent points, i.e.,
pre-training on open geo-spatial and map data is more effective than
using a generic model pre-trained on random web images from Pascal
VOC.
While pre-training is nowadays a standard practice, we go one step
further and pre-train \textit{with aerial images and the correct set
  of output labels}, generated automatically from free map data.

Compared to the gold standard baseline {\bf II} the
performance is $\approx$4 percent points lower ($0.837$
vs.\ $0.874$). In other words, fine-tuning with a limited quantity of
problem-specific high-accuracy labels compensates a large portion
($\approx\,$65\%) of the loss between experiments {\bf
  II} and {\bf IV}, with only $15\,$\% of the labeling
effort.
Relative to {\bf II}, buildings degrade most ($0.863$
vs.\ $0.913$). This can possibly be attributed to the different
appearance of buildings due to different ortho-rectification. Recall
that Google images were rectified with a DTM and are thus
geometrically distorted, with partially visible facades. It seems that
fine-tuning with only three true orthophotos ($<$100 buildings) is not
sufficient to fully adjust the model to the different projection.

Pushing the ``open training data'' philosophy to the extreme, one
could ask whether project-specific training is necessary at all. Maybe
the model learned from open data generalizes even to radiometrically
different images of comparable GSD?
We do not expect this to work, but as a sanity check for a ``generic,
global'' semantic segmentation model we perform a further experiment,
where we avoid domain-adaption altogether. The FCN is trained on all
Google aerial images plus OSM ground truth (Chicago, Paris, Zurich,
Berlin, Potsdam), and then used to predict from the ISPRS images.
This achieves significantly worse results ($0.645$ \(F_1\)-score). A
small set of images with similar radiometry is needed to adapt
the classifier to the sensor properties and lighting conditions of the
test set.

Finally, we respond to the questions we raised at the beginning of this section. A general consensus is that \textit{complete substitution} of manually acquired labels achieves acceptable results. Semantic segmentation of overhead images can indeed be learned from OSM maps without any manual labeling effort albeit at the cost of reduced segmentation accuracy. 
\textit{Augmentation} of manually labeled training data at very large scale reaches the best overall results. Pre-training with large-scale OSM data and publicly available images does improve segmentation of a project-specific data set of independently acquired images and labels (although only by a small margin in this case). An interesting result is that large-scale pre-training on (inaccurate) data increases recall significantly whereas precision slightly drops (compare \textbf{II} and \textbf{V} in Table~\ref{tab:potsdam_results}). 
\textit{Partial substitution} of manually labeled training data with large-scale but inaccurate, publicly available data works very well and seems to be a good trade-off between manual labeling effort and segmentation performance. Indeed, pre-training with large-scale OSM labels \textit{can} replace the vast majority of manual labels. A generic model learned from OSM data adapts very well to a specific location and data source with only little dedicated training data.



\section{CONCLUSIONS AND OUTLOOK}\label{CONCLUSION}

Traditionally, semantic segmentation of aerial and satellite images
crucially relies on manually labelled images as training data. Generating such
training data for a new project is costly and time-consuming, and
presents a bottleneck for automatic image analysis.
The advent of powerful, but data-hungry deep learning methods
aggravates that situation.
Here, we have explored a possible solution, namely to exploit existing
data, in our case open image and map data from the internet for
supervised learning with deep CNNs. Such training data is available in
much larger quantities, but ``weaker'' in the sense that the images are
not representative of the test images' radiometry, and labels
automatically generated from external maps are noisier than dedicated
ground truth annotations.

We have conducted a number of experiments that validate our hypothesis 
stated in the introduction of this paper:
(i) the sheer volume of training data can (largely) compensate for lower accuracy, 
(ii) the large variety present in very large training sets spanning multiple different cities 
does improve the classifier's ability to generalize to new, unseen locations (see predictions on Tokyo, Fig.~\ref{fig:results_tokyo}),
(iii) even if high-quality training data is available, the large volume of additional training data improves classification,
(iv) large-scale (but low-accuracy) training data allows substitution of the large majority (85\% in our case) of the manually annotated high-quality data.

In summary, we can state that weakly labelled training data, when used at large scale, nevertheless 
significantly improves segmentation performance, and improves generalization ability of the models.
We found that even training only on open data, without any manual
labelling, achieves reasonable (albeit far from optimal) results, if
the train/test images are from the same source.
Large-scale pre-training with OSM labels and publicly available aerial
images, followed by domain adaptation to tune to the images at hand,
significantly benefits semantic segmentation and should be used as standard
practice, as long as suitable images and map data are available.

Online map data, as used in our study, is presently limited to RGB
orthophotos with unknown radiometric calibration and street map data
for navigation purposes.
But we are convinced that comparable training databases can be
generated automatically for many problems of interest on the basis of
the image and map archives of mapping agencies and satellite data
providers.
In fact, we are already observing a trend towards free and open data
(e.g., the Landsat and MODIS archives, open geodata initiatives from
several national mapping agencies, etc.).

At first glance, it seems that object classes with complex contextual
relations, like our \textit{road} class, benefit most from more
training data. This intuitively makes sense, because more data is
needed to learn complex long-range layout constraints from data, but
more research is needed to verify and understand the effects in
detail.
Moreover, more studies are needed with different class nomenclatures,
and more diverse datasets, covering different object scales and image
resolutions.
A visionary goal would be a large, free, publicly available ``model
zoo'' of pre-trained classifiers for the most important remote sensing
applications, from which users world-wide can download suitable models
and either apply them directly to their region of interest, or use
them as initialization for their own training.

{
  \bibliographystyle{IEEEtran} 
		\bibliography{TGRS2016_mapgeneration} 
}

\end{document}